\documentclass[a4paper]{article}

\usepackage{INTERSPEECH2019}

\usepackage{enumitem}
\usepackage[dvipsnames]{xcolor}
\usepackage{multirow}
\usepackage{array}

\usepackage{setspace,url}
\PassOptionsToPackage{hyphens}{url}
\usepackage[bookmarks=false,hidelinks]{hyperref}
\usepackage{lineno,multirow,cite}
\usepackage{authblk}
\usepackage{epsfig,amssymb}
\usepackage[utf8]{inputenc}
\usepackage{amsmath,graphicx}
\usepackage{amssymb,epsfig,url,mathrsfs} \usepackage{algorithm,algorithmic}
\usepackage{xcolor}
\usepackage{multirow}
\usepackage{multicol}
\usepackage{enumitem}

\usepackage{numprint}
\npthousandsep{,}

\usepackage{graphicx}
\usepackage{subfig}
\usepackage{array, makecell}
\usepackage{verbatim}

\usepackage[utf8]{inputenc}
\usepackage[T1]{fontenc}
\usepackage[english]{babel}
\usepackage{tabularx}  % for 'tabularx' environment and 'X' column type
\usepackage{ragged2e}  % for '\RaggedRight' macro (allows hyphenation)
\newcolumntype{Y}{>{\RaggedRight\arraybackslash}X} 
\usepackage{booktabs}  % for \toprule, \midrule, and \bottomrule macros 
\title{Introducing the VoicePrivacy Initiative}

\name{N. Tomashenko$^1$, 
B.~M.~L. Srivastava$^2$,
X. Wang$^3$, 
E. Vincent$^4$,
A. Nautsch$^5$,
J. Yamagishi$^{3,6}$,
N.~Evans$^5$,
J. Patino$^5$,
J.-F. Bonastre$^1$,
P.-G. Noé$^1$,
M. Todisco$^5$}

\address{
  $^1$LIA, University of Avignon, France \ \ $^2$Inria, France \ \ $^3$NII, Tokyo, Japan \ \ $^4$Université de Lorraine, CNRS, Inria, LORIA, France \ \ $^5$EURECOM, France \ \ $^6$University of Edinburgh, UK}
  
\email{organisers@lists.voiceprivacychallenge.org}

\begin{document}

\maketitle
\begin{abstract}
The VoicePrivacy initiative aims to promote the development of privacy preservation tools for speech technology by gathering a new community to define the tasks of interest and the evaluation methodology, and benchmarking solutions through a series of challenges. In this paper, we formulate the voice anonymization task selected for the VoicePrivacy 2020 Challenge and describe the datasets used for system development and evaluation. We also present the attack models and the associated objective and subjective evaluation metrics. We introduce two anonymization baselines and report objective evaluation results. 
\end{abstract}


\section{Introduction}\label{sec:intro}
Recent years have seen mounting calls for the preservation of privacy when treating or storing personal data. This is not least the result of the European general data protection regulation (GDPR). While there is no legal definition of privacy \cite{nautsch2019gdpr}, speech data encapsulates a wealth of personal information that can be revealed by listening or by automated systems \cite{Nautsch-PreservingPrivacySpeech-CSL-2019}. This includes, e.g., age, gender, ethnic origin, geographical background, health or emotional state, political orientations, and religious beliefs, among others \cite[p.~62]{COMPRISE_D5.1}. In addition, speaker recognition systems can reveal the speaker's identity. It is thus of no surprise that efforts to develop privacy preservation solutions for speech technology are starting to emerge. The VoicePrivacy initiative aims to gather a new community to define the tasks of interest and the evaluation methodology, and to benchmark these solutions through a series of challenges.

Current methods fall into four categories: deletion, encryption, distributed learning, and anonymization. Deletion methods \cite{cohen2019voice,gontier2020privacy} are meant for ambient sound analysis. They delete or obfuscate any overlapping speech to the point where no information about it can be recovered. Encryption methods \cite{pathak2013privacy,smaragdis2007framework} such as fully homomorphic encryption \cite{zhang2019encrypted} and secure multiparty computation \cite{brasser2018voiceguard}, support computation upon data in the encrypted domain. They incur significant increases in computational complexity, which require special hardware. Decentralized or federated learning methods aim to learn models from distributed data without accessing it directly \cite{leroy2019federated}. The derived data used for learning (e.g., model gradients) may still leak information about the original data, however \cite{geiping2020inverting}.

Anonymization refers to the goal of suppressing personally identifiable attributes of the speech signal, leaving all other attributes intact\footnote{In the legal community, the term ``anonymization'' means that this goal has been achieved. Here, it refers to the task to be addressed, even when the method being evaluated has failed. We expect the VoicePrivacy initiative to lead to the definition of new, unambiguous terms.}. Past and recent attempts have focused on noise addition \cite{hashimoto2016privacy}, speech transformation \cite{qian2017voicemask}, voice conversion \cite{jin2009speaker,pobar2014online,bahmaninezhad2018convolutional,magarinos2017reversible}, speech synthesis \cite{fang2019speaker,han2020voice}, or adversarial learning \cite{srivastava2019privacy}. In contrast to the above categories of methods, anonymization appears to be more flexible since it can selectively suppress or retain certain attributes and it can easily be integrated within existing systems. Despite the appeal of anonymization and the urgency to address privacy concerns, a formal definition of anonymization and attacks against it is missing. Furthermore, the level of anonymization offered by existing solutions is unclear and not meaningful because there are no common datasets, protocols and metrics. 

For these reasons, the VoicePrivacy 2020 Challenge focuses on the task of speech anonymization. This paper is intended as a general reference about the Challenge for researchers, engineers and privacy professionals. Details for participants are provided in the evaluation plan \cite{tomashenkovoiceprivacy} and on the challenge website\footnote{\url{https://www.voiceprivacychallenge.org/}}.

The paper is structured as follows. The anonymization task and the attack models, the datasets, and the metrics are described in Sections~\ref{sec:task}, \ref{sec:data}, and \ref{sec:perf}, respectively. The two baseline systems and the corresponding objective evaluation results are presented in Section~\ref{sec:baseline}. We conclude in Section \ref{sec:conclusions}.

\section{Anonymization task and attack models}
\label{sec:task}
Privacy preservation is formulated as a game between \emph{users} who publish some data and \emph{attackers} who access this data or data derived from it and wish to infer information about the users \cite{qian2018towards,srivastava2019evaluating}. To protect their privacy, the users publish data that contain as little personal information as possible while allowing one or more downstream goals to be achieved. To infer personal information, the attackers may use additional prior knowledge.

Focusing on speech data, a given privacy preservation scenario is specified by:
(i)~the nature of the data: waveform, features, etc., (ii)~the information seen as personal: speaker identity, traits, spoken contents, etc., (iii)~the downstream goal(s): human communication, automated processing, model training, etc., (iv)~
the data accessed by the attackers: one or more utterances, derived data or model, etc., (v)~the attackers' prior knowledge: previously published data, privacy preservation method applied, etc. Different specifications lead to different privacy preservation methods from the users' point of view and different attacks from the attackers' point of view.

\subsection{Privacy preservation scenario}
VoicePrivacy 2020 considers the following scenario, where the terms \textsl{``user''} and \textsl{``speaker''} are used interchangeably.
Speakers want to hide their identity while still allowing all other downstream goals to be achieved. Attackers have access to one or more utterances and want to identify the speakers.

\subsection{Anonymization task}
\label{subsec:user_goals}
To hide his/her identity, each speaker passes his/her utterances through an anonymization system. The resulting anonymized utterances are referred to as \emph{trial} data. They sound as if they had been uttered by another speaker called \emph{pseudo-speaker}, which may be an artificial voice not corresponding to any real speaker.

The task of challenge participants is to design this anonymization system. In order to allow all downstream goals to be achieved, this system should: (a)~output a speech waveform, (b)~hide speaker identity as much as possible, (c)~distort other speech characteristics as little as possible, (d)~ensure that all trial utterances from a given speaker appear to be uttered by the same pseudo-speaker, while trial utterances from different speakers appear to be uttered by different pseudo-speakers\footnote{This is akin to ``pseudonymization'', which replaces each user's identifiers by a unique key. We do not use this term here, since it often refers to the distinct case when the identifiers are tabular data and the data controller stores the correspondence table linking users and keys.}.

Requirement (c) is assessed via \emph{utility} metrics: automatic speech recognition (ASR) decoding error rate using a model trained on \emph{original}, i.e., unprocessed data and subjective speech intelligibility and naturalness (see Section \ref{sec:perf}). Requirement (d) and additional downstream goals including ASR training will be assessed in a post-evaluation phase (see Section \ref{sec:conclusions}).

\subsection{Attack models}
\label{subsec:attack_model}
The attackers have access to: (a) one or more anonymized trial utterances, (b) possibly, original or anonymized \emph{enrollment} utterances for each speaker. They do not have access to the anonymization system applied by the user. The protection of personal information is assessed via \emph{privacy} metrics, including objective speaker verifiability and subjective speaker verifiability and linkability. These metrics assume different attack models.

The objective speaker verifiability metrics assume that the attackers have access to a single anonymized trial utterance and several enrollment utterances. Two sets of metrics are used for original vs.\ anonymized enrollment data (see Section \ref{sec:perf_objective}). In the latter case, we assume that the trial and enrollment utterances of a given speaker have been anonymized using the same system, but the corresponding pseudo-speakers are different.

The subjective speaker verifiability metric (Section \ref{sec:perf_subjective}) assumes that the attackers have access to a single anonymized trial utterance and a single original enrollment utterance. Finally, the subjective speaker linkability metric (Section \ref{sec:perf_subjective}) assumes that the attackers have access to several anonymized trial utterances.

\section{Datasets}\label{sec:data}
Several publicly available corpora are used for the training, development and evaluation of speaker anonymization systems.

\subsection{Training set}\label{sec:train-data}
The training set comprises the \numprint{2800}~h \textit{VoxCeleb-1,2} speaker verification corpus \cite{nagrani2017voxceleb,chung2018voxceleb2} and 600~h subsets of the \textit{{LibriSpeech}} \cite{panayotov2015librispeech} and \textit{LibriTTS} \cite{zen2019libritts} corpora, which were initially designed for ASR and speech synthesis, respectively. The selected subsets are detailed in Table~\ref{tab:data} (top).

\begin{table}[htbp]
  \caption{Number of speakers and utterances in the VoicePrivacy 2020 training, development, and evaluation sets.}\label{tab:data}
\renewcommand{\tabcolsep}{0.09cm}
  \centering
  \footnotesize
  \begin{tabular}{|l|l|l|r|r|r|r|}
\hline
 \multicolumn{3}{|l|}{\textbf{Subset}} &  \textbf{Female} & \textbf{Male} & \textbf{Total} & \textbf{\#Utter.}  \\ \hline \hline
% train
\multirow{5}{*}{\rotatebox{90}{Training~}} & \multicolumn{2}{l|}{VoxCeleb-1,2} & \numprint{2912} & \numprint{4451} & \numprint{7363} & \numprint{1281762} \\ \cline{2-7}
& \multicolumn{2}{l|}{LibriSpeech train-clean-100} & 125 & 126 & 251	& \numprint{28539} \\ \cline{2-7}
& \multicolumn{2}{l|}{LibriSpeech train-other-500} & 564 & 602 & \numprint{1166} & \numprint{148688}	\\ \cline{2-7}
& \multicolumn{2}{l|}{LibriTTS train-clean-100} & 123 & 124 & 247 & \numprint{33236} \\ \cline{2-7}
& \multicolumn{2}{l|}{LibriTTS train-other-500} & 560 & 600 & \numprint{1160} & \numprint{205044} \\ \hline\hline
% devel
\multirow{5}{*}{\rotatebox{90}{Development~}} & LibriSpeech & Enrollment & 15 & 14 & 29 & 343\\ \cline{3-7}
& dev-clean & Trial & 20 & 20 & 40 & \numprint{1978}\\ \cline{2-7}
& & Enrollment & & & & 600 \\  \cline{3-3}\cline{7-7}
& VCTK-dev & Trial (common) & 15 & 15 & 30 & 695\\  \cline{3-3}\cline{7-7}
& & Trial (different) & & & & \numprint{10677} \\  \hline\hline
% eval
\multirow{5}{*}{\rotatebox{90}{Evaluation~}} & LibriSpeech & Enrollment & 16 & 13 & 29 & 438\\ \cline{3-7}
& test-clean & Trial & 20 & 20 & 40 & \numprint{1496}\\ \cline{2-7}
& & Enrollment & & & & 600 \\  \cline{3-3}\cline{7-7}
& VCTK-test & Trial (common) & 15 & 15 & 30 & 70 \\ \cline{3-3}\cline{7-7}
& & Trial (different) & & & & \numprint{10748} \\ \hline
\end{tabular}
\end{table}
\normalsize

\subsection{Development set}\label{sec:data-dev}
The development set comprises \textit{LibriSpeech dev-clean} and a subset of the VCTK corpus \cite{yamagishi2019cstr} denoted as \textit{VCTK-dev} (see Table~\ref{tab:data}, middle). With the above attack models in mind, we split them into trial and enrollment subsets. For \textit{LibriSpeech dev-clean}, the speakers in the enrollment set are a subset of those in the trial set. For \textit{VCTK-dev}, we use the same speakers for enrollment and trial and we consider two trial subsets, denoted as \textit{common} and \textit{different}. The \textit{common} trial subset is composed of utterances $\#1-24$ in the VCTK corpus that are identical for all speakers. This is meant for subjective evaluation of speaker verifiability/linkability in a text-dependent manner. The enrollment and \textit{different} trial subsets are composed of distinct utterances for all speakers.

\subsection{Evaluation set}
Similarly, the evaluation set comprises \textit{LibriSpeech test-clean} and a subset of VCTK called \textit{VCTK-test} (see Table~\ref{tab:data}, bottom).

\section{Utility and privacy metrics}
\label{sec:perf}

Following the attack models in Section \ref{subsec:attack_model}, we consider objective and subjective privacy metrics to assess anonymization performance in terms of speaker verifiability and linkability. We also propose objective and subjective utility metrics to assess whether the requirements in Section \ref{subsec:user_goals} are fulfilled.

\subsection{Objective metrics}
\label{sec:perf_objective}
For objective evaluation, we train two systems to assess speaker verifiability and ASR decoding error. The first system denoted $ASV_\text{eval}$ is an automatic speaker verification (ASV) system, which produces log-likelihood ratio (LLR) scores. The second system denoted $ASR_\text{eval}$ is an ASR system which outputs a word error rate (WER). Both are trained on \textit{LibriSpeech train-clean-360} using Kaldi \cite{povey2011kaldi}.

\subsubsection{Objective speaker verifiability}\label{sec:asv-eval} 

The $ASV_\text{eval}$ system for speaker verifiability evaluation relies on x-vector speaker embeddings and probabilistic linear discriminant analysis (PLDA) \cite{snyder2018x}. Three metrics are computed: the equal error rate (EER) and the LLR-based costs $C_\text{llr}$ and $C^\text{min}_\text{llr}$. Denoting by $P_\text{fa}(\theta)$ and $P_\text{miss}(\theta)$ the false alarm and miss rates at threshold~$\theta$, the EER corresponds to the threshold $\theta_\text{EER}$ at which the two detection error rates are equal, i.e., $\text{EER}=P_\text{fa}(\theta_\text{EER})=P_\text{miss}(\theta_\text{EER})$. $C_\text{llr}$ is computed from PLDA scores as defined in \cite{brummer2006application,ramos2008cross}.
It can be decomposed into a discrimination loss ($C^\text{min}_\text{llr}$) and a calibration loss ($C_\text{llr}-C^\text{min}_\text{llr}$). $C^\text{min}_\text{llr}$ is estimated by optimal calibration using monotonic transformation of the scores to their empirical LLR values.

\begin{figure}[htbp]
\centering\includegraphics[width=79mm]{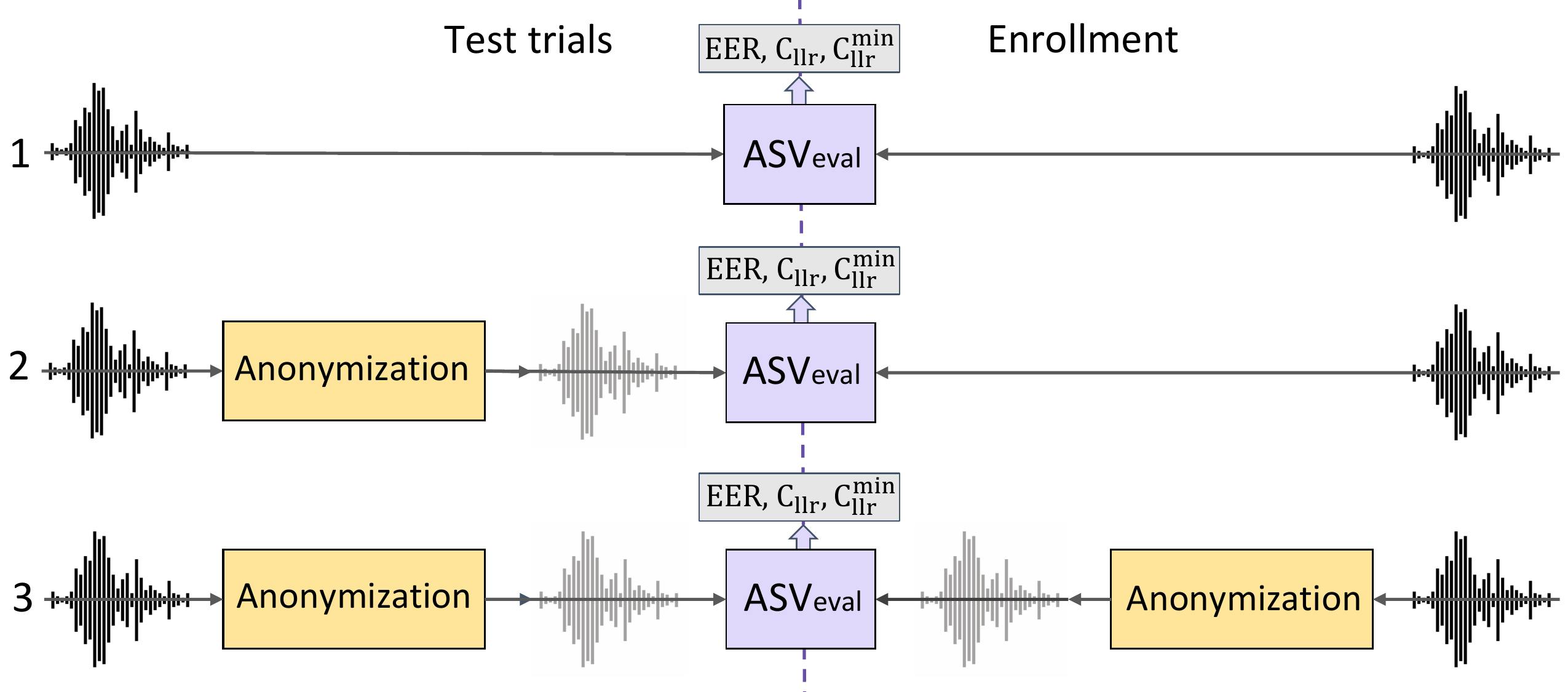}
\caption{ASV evaluation.}
\label{fig:asv-eval}
\end{figure}

As shown in Fig.~\ref{fig:asv-eval}, these metrics are computed and compared for: (1)~original trial and enrollment data, (2)~anomymized trial data and original enrollment data, (3)~anomymized trial and enrollment data. The number of target and impostor trials is given in Table \ref{tab:trials}.

\begin{table}[htbp]
  \caption{Number of speaker verification trials.}\label{tab:trials}
  \renewcommand{\tabcolsep}{0.13cm}
  \centering
  \footnotesize
  \begin{tabular}{|l|l|l|r|r|r|}
\hline
 \multicolumn{2}{|l|}{\textbf{Subset}} & \textbf{Trials} &  \textbf{Female} & \textbf{Male} & \textbf{Total}  \\ \hline \hline
% dev
\multirow{6}{*}{\rotatebox{90}{Development~}} & LibriSpeech & Target & 704 & 644 & \numprint{1348} \\ \cline{3-6}
 & dev-clean & Impostor	& \numprint{14566} & \numprint{12796} &	\numprint{27362} \\ \cline{2-6}
 & \multirow{4}{*}{VCTK-dev} & Target (common) & \numprint{344} & \numprint{351} & \numprint{695} \\ \cline{3-6}
 & & Target (different) & \numprint{1781}	& \numprint{2015} & \numprint{3796} \\  \cline{3-6}
 & & Impostor (common) & \numprint{4810} &	\numprint{4911} & \numprint{9721} \\ \cline{3-6}
 & & Impostor (different) & \numprint{13219} & \numprint{12985} & \numprint{26204} \\ \hline\hline
% eval
\multirow{6}{*}{\rotatebox{90}{Evaluation~}} & LibriSpeech & Target & 548 & 449	& \numprint{997} \\ \cline{3-6}
 & test-clean & Impostor & \numprint{11196} & \numprint{9457} &	\numprint{20653} \\ \cline{2-6}
 & \multirow{4}{*}{VCTK-test} & Target (common) & \numprint{346} & \numprint{354} & \numprint{700} \\ \cline{3-6}
 & & Target (different) & \numprint{1944} & \numprint{1742} & \numprint{3686} \\  \cline{3-6}
 & & Impostor (common) & \numprint{4838} & \numprint{4952} & \numprint{9790} \\ \cline{3-6}
 & & Impostor (different) & \numprint{13056} &	\numprint{13258} &\numprint{26314} \\ \hline
  \end{tabular}
\end{table}
\normalsize

\subsubsection{ASR decoding error}\label{sec:wer}
\begin{figure}[htbp]
\centering\includegraphics[width=47mm]{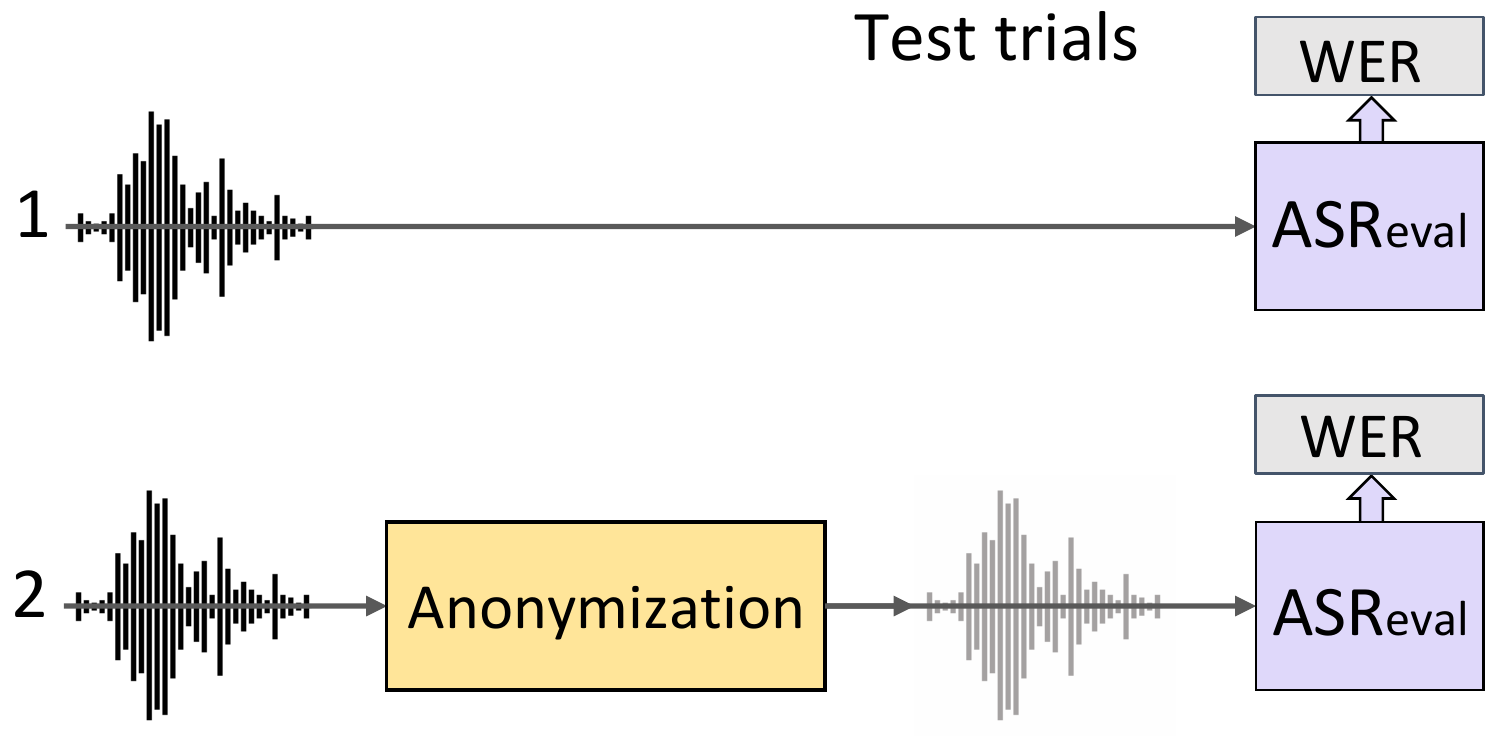}
\caption{ASR decoding evaluation.}
\label{fig:asr-eval}
\end{figure}

$ASR_\text{eval}$ is based on the state-of-the-art Kaldi recipe for LibriSpeech involving a factorized time delay neural network (TDNN-F) acoustic model (AM) \cite{povey2018semi,peddinti2015time} and a trigram language model. As shown in Fig.~\ref{fig:asr-eval}, the (1) original and (2) anonymized trial data is decoded using the provided pretrained $ASR_\text{eval}$ model and the corresponding WERs are calculated.

\subsection{Subjective metrics}\label{sec:subj-metr}
\label{sec:perf_subjective}
Subjective metrics include speaker verifiability, speaker linkability, speech intelligibility, and speech naturalness. They will be evaluated using listening tests carried out by the organizers.

\subsubsection{Subjective speaker verifiability}
To evaluate subjective speaker verifiability, listeners are given pairs of one anonymized trial utterance and one distinct original enrollment utterance of the same speaker. Following \cite{lorenzo2018voice}, they are instructed to imagine a scenario in which the anonymized sample is from an incoming telephone call, and to rate the similarity between the voice and the original voice using a scale of 1 to 10, where 1 denotes `different speakers' and 10 denotes `the same speaker' with highest confidence. The performance of each anonymization system will be visualized through detection error tradeoff (DET) curves.

\subsubsection{Subjective speaker linkability}
The second subjective metric assesses speaker linkability, i.e., the ability to cluster several utterances into speakers. Listeners are asked to place a set of anonymized trial utterances from different speakers in a 1- or 2-dimensional space according to speaker similarity. This relies on a graphical interface, where each utterance is represented as a point in space and the distance between two points expresses subjective speaker dissimilarity.

\subsubsection{Subjective speech intelligibility}
Listeners are also asked to rate the intelligibility of individual samples (anonymized trial utterances or original enrollment utterances) on a scale from 1 (totally unintelligible) to 10 (totally intelligible). The results can be visualized through DET curves.

\subsubsection{Subjective speech naturalness}
Finally, the naturalness of the anonymized speech will be evaluated on a scale from 1 (totally unnatural) to 10 (totally natural).

\begin{table*}[htp]
  \caption{Speaker verifiability achieved by the pretrained \emph{$ASV_\text{eval}$} model. The primary baseline is used for anonymization.}\label{tab:asv-results}
  \renewcommand{\tabcolsep}{0.24cm}
  \centering
  \footnotesize{
  \begin{tabular}{|c|c|c|c|r|r|r|r|r|r|}
\hline
\multirow{2}{*}{\textbf{Dataset}} & \multirow{2}{*}{\textbf{Gender}} & \multicolumn{2}{c|}{\textbf{Anonymization}} & \multicolumn{3}{c|}{\textbf{Development}} & \multicolumn{3}{c|}{\textbf{Test}}\\ \cline{3-10}
 & & \textbf{Enroll} & \textbf{Trial} & \textbf{EER (\%)} & $\mathbf{C}_{\textbf{llr}}^{\textbf{min}}$ & $\mathbf{C}_{\textbf{llr}}$ & \textbf{EER (\%)} & $\mathbf{C}_{\textbf{llr}}^{\textbf{min}}$ & $\mathbf{C}_{\textbf{llr}}$ \\ \hline \hline
\multirow{6}{*}{LibriSpeech} & \multirow{3}{*}{Female} & \multirow{2}{*}{original} & original & 8.67 & 0.304 & 42.86 & 7.66 & 0.183 & 26.79\\  \cline{4-10}
 & & & \multirow{2}{*}{anonymized} & 50.14 & 0.996 & 144.11  & 47.26 & 0.995 & 151.82\\  \cline{3-3}\cline{5-10}
 & & anonymized & & 36.79 & 0.894 & 16.35 & 32.12 & 0.839 & 16.27\\  \cline{2-10}
 & \multirow{3}{*}{Male} & \multirow{2}{*}{original} & original & 1.24 & 0.034 & 14.25 & 1.11 & 0.041 & 15.30\\  \cline{4-10}
 & & & \multirow{2}{*}{anonymized} & 57.76 & 0.999 & 168.99 & 52.12 & 0.999 & 166.66\\  \cline{3-3}\cline{5-10}
 & & anonymized & & 34.16 & 0.867 & 24.72 & 36.75 & 0.903 & 33.93\\ \hline
\multirow{6}{*}{VCTK} & \multirow{3}{*}{Female} & \multirow{2}{*}{original} & original &  2.86 & 0.100 & 1.13  & 4.89 & 0.169 & 1.50\\  \cline{4-10}
\multirow{6}{*}{(different)} & & & \multirow{2}{*}{anonymized} & 49.97 & 0.989 & 166.03 & 48.05 & 0.998 & 146.93\\  \cline{3-3}\cline{5-10}
 & & anonymized & &  26.11 & 0.760 & 8.41 & 31.74 & 0.847 & 11.53\\  \cline{2-10}
 & \multirow{3}{*}{Male} & \multirow{2}{*}{original} & original & 1.44 & 0.052 & 1.16  & 2.07 & 0.072 & 1.82\\  \cline{4-10}
 & & & \multirow{2}{*}{anonymized} & 53.95 & 1.000 & 167.51 & 53.85 & 1.000 & 167.82\\  \cline{3-3}\cline{5-10}
 & & anonymized & & 30.92 & 0.839 & 23.80 & 30.94 & 0.834 & 23.84\\ \hline 
  \end{tabular}  }
\end{table*}

\begin{table}[htp]
  \caption{ASR decoding error achieved by the pretrained \emph{$ASR_\text{eval}$} model. The primary baseline is used.}\label{tab:asr-results}
  \centering
  \footnotesize
  \renewcommand{\tabcolsep}{0.11cm}
  \begin{tabular}{|c|c|r|r|}
\hline
\textbf{Dataset} & \textbf{Anonymization} & \textbf{Dev.\ WER (\%)} & \textbf{Test WER (\%)} \\ \hline \hline
\multirow{2}{*}{LibriSpeech} & original & 3.83 & 4.15\\ \cline{2-4} 
 & anonymized & 6.39 & 6.73\\ \hline
VCTK & original & 10.79 & 12.82\\ \cline{2-4} 
(comm.+diff.) & anonymized & 15.38 & 15.23\\ \hline  
  \end{tabular}
\end{table}
\normalsize

\section{Baseline software and results}\label{sec:baseline}
Two anonymization baselines are provided.\footnote{\url{https://github.com/Voice-Privacy-Challenge/Voice-Privacy-Challenge-2020}} We briefly introduce them and report the corresponding objective results below.

\subsection{Anonymization baselines}
\begin{figure}[bp]
\centering\includegraphics[width=80mm]{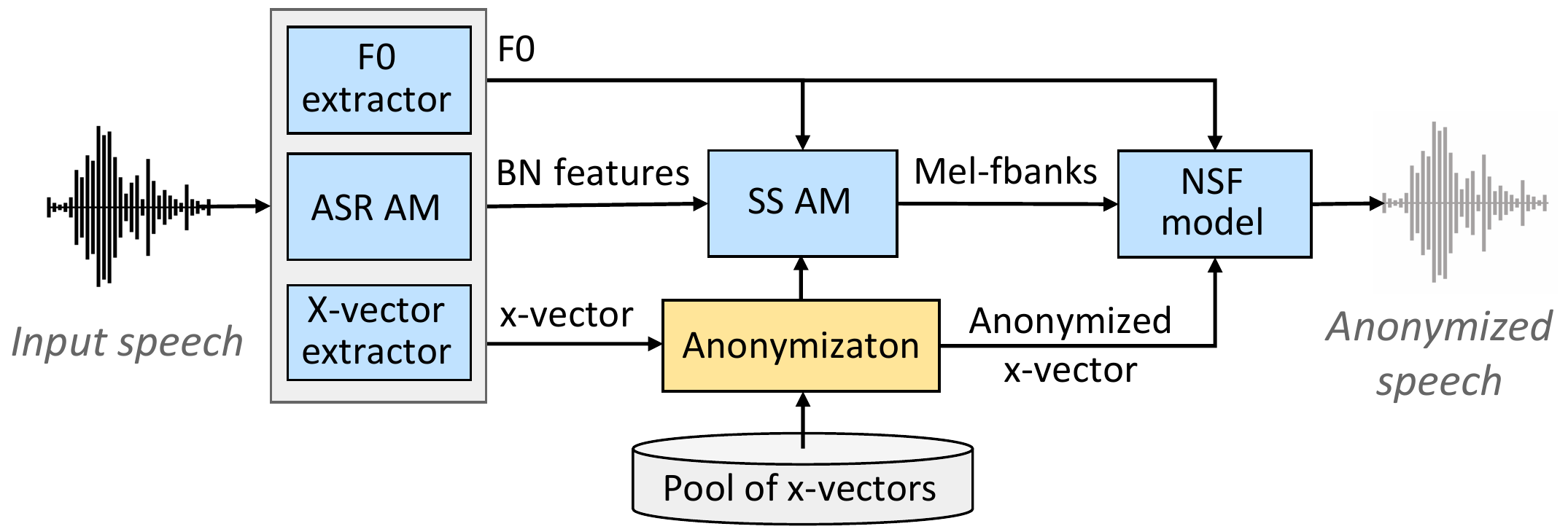}
\caption{Primary baseline anonymization system.}
\label{fig:baseline}
\end{figure}

The primary baseline shown in Fig.~\ref{fig:baseline} is inspired from \cite{fang2019speaker} and comprises three steps: (1) extraction of x-vector \cite{snyder2018x}, pitch (F0) and bottleneck (BN) features; (2) x-vector anonymization; (3) speech synthesis (SS) from the anonymized x-vector and the original F0+BN features. In \textsl{Step 1}, 256-dimensional BN features encoding spoken content are extracted using a TDNN-F ASR AM trained on \textit{LibriSpeech train-clean-100} and \textit{train-other-500} using Kaldi. A 512-dimensional x-vector encoding the speaker is extracted using a TDNN trained on \textit{VoxCeleb-1,2} with Kaldi. In \textsl{Step 2}, for every source x-vector, an anonymized x-vector is computed by finding the $N$ farthest x-vectors in an external pool (\textit{LibriTTS train-other-500}) according to the PLDA distance, and by averaging $N^*$ randomly selected vectors among them\footnote{In the baseline, we use $N=200$ and $N^*=100$.}. In \textsl{Step 3}, an SS AM generates Mel-filterbank features given the anonymized x-vector and the F0+BN features, and a neural source-filter (NSF) waveform model \cite{wang2019neural} outputs a speech signal given the anonymized x-vector, the F0, and the generated Mel-filterbank features. The SS AM and NSF models are both trained on \textit{LibriTTS train-clean-100}. See \cite{tomashenkovoiceprivacy,srivastava2020baseline} for further details.

The secondary baseline is a simpler, formant-shifting approach provided as additional inspiration \cite{EURECOM+6190}.

\subsection{Objective evaluation results}
Table~\ref{tab:asv-results} reports the values of objective speaker verifiability metrics obtained before/after anonymization with the primary baseline.\footnote{Results on VCTK (common) are omitted due to space constraints.} The EER and $C_\text{llr}^\text{min}$ metrics behave similarly, while interpretation of $C_\text{llr}$ is more challenging due to non-calibration\footnote{In particular, $C_\text{llr}>1$ is not a problem, since we care more about discrimination metrics than score calibration metrics in the first edition.}. We hence focus on the EER below. On all datasets, anonymization of the trial data greatly increases the EER. This shows that the anonymization baseline effectively increases the users' privacy. 
The EER estimated with original enrollment data (47 to 58\%), which is comparable to or above the chance value (50\%), suggests that full anonymization has been achieved. However, anonymized enrollment data result in a much lower EER (26 to 37\%), which suggests that F0+BN features retain some information about the original speaker.
If the attackers have access to such enrollment data, they will be able to re-identify users almost half of the time. Note also that the EER is larger for females than males on average. This further demonstrates that failing to define the attack model or assuming a naive attack model leads to a greatly overestimated sense of privacy~\cite{srivastava2019evaluating}.

Table~\ref{tab:asr-results} reports the WER achieved before/after anonymization with the primary baseline. 
While the absolute WER stays below 7\% on LibriSpeech and 16\% on VCTK, anonymization incurs a large WER increase of 19 to 67\% relative.

The results achieved by the secondary baseline are inferior and detailed in \cite{tomashenkovoiceprivacy}. Overall, there is substantial potential for challenge participants to improve over the two baselines.

\section{Conclusions}
\label{sec:conclusions}
The VoicePrivacy initiative aims to promote the development of private-by-design speech technology. Our initial event, the VoicePrivacy 2020 Challenge, provides a complete evaluation protocol for voice anonymization systems. We formulated the voice anonymization task as a game between users and attackers, and highlighted three possible attack models. We also designed suitable datasets and evaluation metrics, and we released two open-source baseline voice anonymization systems. Future work includes evaluating and comparing the participants' systems using objective and subjective metrics, computing alternative objective metrics relating to, e.g., requirement (d) in Section \ref{subsec:user_goals}, and drawing initial conclusions regarding the best anonymization strategies for a given attack model. A revised, stronger evaluation protocol is also expected as an outcome.

In this regard, it is essential to realize that the users' downstream goals and the attack models listed above are not exhaustive. For instance, beyond ASR decoding, anonymization is extremely useful in the context of anonymized data collection for ASR training \cite{srivastava2019privacy}. It is also known that the EER becomes lower when the attackers generate anonymized training data and retrains $ASV_\text{eval}$ on this data \cite{srivastava2019evaluating}. In order to assess these aspects, we will ask volunteer participants to share additional data with us and run additional experiments in a post-evaluation phase.

\section{Acknowledgment}
VoicePrivacy was born at the crossroads of projects VoicePersonae, COMPRISE (\url{https://www.compriseh2020.eu/}), and DEEP-PRIVACY. Project HARPOCRATES was designed specifically to support it. The authors acknowledge support by ANR, JST, and the European Union's Horizon 2020 Research and Innovation Program, and they would like to thank Md Sahidullah and Fuming Fang. Experiments presented in this paper were partially carried out using the Grid'5000 testbed, supported by a scientific interest group hosted by Inria and including CNRS, RENATER and several Universities as well as other organizations (see \url{https://www.grid5000.fr}).

\bibliographystyle{IEEEtran}
\bibliography{mybib}

\end{document}